\documentclass[conference]{IEEEtran}
\IEEEoverridecommandlockouts
\usepackage{cite}
\usepackage{amsmath,amssymb,amsfonts}
\usepackage{algorithmic}
\usepackage{graphicx}
\usepackage{textcomp}
\usepackage{xcolor}
\usepackage{multirow}
\begin{document}

\title{A Modular Multi-stage Lightweight Graph Transformer Network for Human Pose and Shape Estimation from 2D Human Pose}

\author{\IEEEauthorblockN{Ayman Ali, Ekkasit Pinyoanuntapong, Pu Wang, Mohsen Dorodchi}
\IEEEauthorblockA{\textit{College of Computing and Informatics} \\
\textit{University of North Carolina at Charlotte}\\
Charlotte, United States \\
aali26, epinyoan, pwang13, mdorodch@uncc.edu}
}
\graphicspath{{images/}}
\maketitle

\begin{abstract}
In this research, we address the challenge faced by existing deep learning-based human mesh reconstruction methods in balancing accuracy and computational efficiency. These methods typically prioritize accuracy, resulting in large network sizes and excessive computational complexity, which may hinder their practical application in real-world scenarios, such as virtual reality systems. To address this issue, we introduce a modular multi-stage lightweight graph-based transformer network for human pose and shape estimation from 2D human pose, a pose-based human mesh reconstruction approach that prioritizes computational efficiency without sacrificing reconstruction accuracy. Our method consists of a 2D-to-3D lifter module that utilizes graph transformers to analyze structured and implicit joint correlations in 2D human poses, and a mesh regression module that combines the extracted pose features with a mesh template to produce the final human mesh parameters.
\end{abstract}

\section{\textbf{Introduction}}
The field of computer vision has witnessed significant advancements with the advent of deep learning models in human pose estimation from monocular images. Specifically, 2D pose estimation involves determining the locations of human joint coordinates within a single image. In recent years, the research community has shown great interest in reconstructing 3D human mesh representations due to their practical applications in human-computer interaction, gaming, and virtual systems. Despite the progress in both 2D and 3D pose estimation and human mesh reconstruction, reconstructing human mesh from a single image remains challenging due to factors such as depth ambiguity, complex backgrounds, challenging postures, and a lack of annotated in-the-wild datasets.

The fundamental process of reconstructing a 3D human mesh involves estimating the pose and shape parameters based on statistical body models, such as SMPL \cite{loper2015smpl} and \cite{romero2022embodied}. The pose parameters $\theta$ define the global root orientation and the relative rotations of human body joints represented in an axis-angle sequence. Moreover, the shape parameters $\beta$ represent the learned linear gender-based body shape derived from the CAESAR dataset \cite{robinette2002civilian} using principal component analysis.

The literature on human mesh reconstruction primarily consists of two paradigms: optimization-based and regression-based approaches. Optimization-based methods minimize the difference between the re-projected 2D pose regressed from the human mesh and the estimated 2D pose. In contrast, regression-based methods directly predict the pose and shape parameters.

While deep learning models have progressed in human mesh reconstruction, the choice of input data has a significant impact on the system's pipeline. Recent advances have adopted an end-to-end pipeline where the input is the scene image, and the output is the predicted mesh. However, this approach can be computationally intensive and does not meet the efficiency requirements of various real-world applications, such as gaming and real-time mesh recovery.

To address the computational overhead, some researchers have proposed using skeleton data as the input, as it is sparse \cite{choi2020pose2mesh}. However, solely relying on skeleton data is insufficient to address the computational intensity. \textit{Zheng et al.} \cite{zheng2022lightweight} proposed a lightweight graph-based transformer architecture to focus on efficiency.

Nevertheless, reconstructing mesh parameters remains complex, as it is influenced by a variety of factors, including 1) Dataset Characteristics where the performance of a model is not significantly impacted by the indoor/outdoor setting or the number of data points, but rather by human pose and shape, camera characteristics, and backbone features \cite{pang2022benchmarking}. Furthermore, a diverse range of these attributes can improve the performance, while occlusion and SMPL fittings can enhance recovery accuracy. 2) Dataset Mix is another factor that affects the complexity of the task. The choice of datasets is critical for the generalizability and accuracy of the model \cite{pang2022benchmarking}. It is imperative to use the same combination of datasets when evaluating and comparing the impact of other factors, such as training algorithms. Comparing two network architectures on different dataset combinations is considered unfair. To establish a robust baseline model, it is recommended to use more challenging datasets and increase their contribution during training. Finally, Noisy annotations, which in case of having high proportions of noisy data samples, can negatively impact model performance, particularly when both SMPL annotations and keypoints are affected. However, slightly noisy SMPL data can still have a positive effect on training.

Our objective in this research is to design a modular and end-to-end capable graph-based transformer network that is capable of estimating the human shape and pose, demonstrating comparable performance with state-of-the-art methods.

Specifically, we propose the following strategies to achieve this goal:
\begin{itemize}
\item Construct a modular, multi-stage pipeline that is capable of end-to-end estimation of human parameters.
\item separate the learning of human pose, shape, and camera parameters to improve accuracy and robustness of the model.
% \item The integration of temporal information to address the issue of inconsistent pose estimation that arises from frame-level estimation of human pose and shape parameters in videos.
\end{itemize}

\section{System Description}

\subsection{Preliminary}
The task of Human Mesh Recovery (HMR) from images without the utilization of auxiliary devices, such as depth sensors, poses a challenge due to various factors such as depth ambiguity, complex backgrounds, and diverse human poses. Typically, deep learning networks are trained to estimate the human pose parameters using a skinned vertex-based parametric human model such as SMPL \cite{loper2015smpl}.

\textit{Kanazawa et al.} proposed a method that minimizes the 2D reprojection loss of joints without relying on paired 2D-to-3D supervision to estimate the pose and shape parameters from an image \cite{kanazawa2018end}. \textit{Kolotouros et al.} proposed a method that iteratively optimizes the estimated parameters to produce a pixel-level alignment with the original image \cite{kocabas2020vibe}. Further, \textit{Kocabas et al.} introduced adversarial training to enhance the quality of the estimated mesh by utilizing the large-scale motion capture dataset AMASS \cite{mahmood2019amass}.

Recent advancements in 2D pose estimation have prompted researchers to explore the potential of using off-the-shelf 2D pose estimation networks, such as HRNet \cite{sun2019deep}, to estimate human mesh parameters. The 2D skeleton estimated from an image is fed to the human mesh parameters estimation network for further processing and analysis.

\textbf{Graph Convolution Networks} (GCN) have received considerable attention in recent years due to their ability to intuitively model data, such as articulated models. In particular, many studies have adopted GCN to model 3D human pose estimation \cite{zou2021modulated,zhao2019semantic}. In the context of human skeletons, joints and bones are mapped to vertices and edges, respectively, thus forming a graph represented mathematically as $\mathcal{G} = {v, \epsilon}$. Inspired by \cite{zheng2022lightweight}, we use GCN to model the 2D pose features, where the output of the GCN layer is expressed as:
\begin{equation}
X^\prime = \sigma(A \times W)
\end{equation}
Where $\sigma(.)$ represents the Gaussian Error Linear Unit (GELU), $A$ represents the adjacency matrix, and $W$ represents the learnable weight matrix.

\textbf{Transformer} 
Transformer \cite{2017attention} has remodeled many deep learning models in computer vision. The ability to capture the global context understanding in the features space led to significant advancements. 
Self-attention can be described as a function to compute the attention matrix given a query matrix \(Q\in \mathbb{R}^{N\times d}\), key matrix \(K\in \mathbb{R}^{N\times d}\) and value matrix \(V\in \mathbb{R}^{N\times d}\) where $N$ represents the number of vectors in the sequence, and \(d\) is the dimension. A scaling factor (\(\frac{1}{\sqrt{d}}\)) is utilized to alleviate the growth of the softmax function's magnitude. The scaled-dot product attention can be expressed as: 
\begin{equation} Attention(Q,K,V)=Softmax(\frac{QK^T}{\sqrt{d_{k}}})V\end{equation}
\textbf{Multi-headed self-attention} utilized parallel scaled-dot attentions to project the queries, keys, and values $h$ times with different linear projections to $dk$, $dk$, and $DV$ dimensions. A concatenation of the $h$ head attention output can be expressed as: 
\begin{equation}\label{eq:msa}
\begin{aligned}
MSA(Q,K,V) = Concat(H_{1}, H_{2},...., H_{h})W_{out} \\
\text{where}\; H_{i} = Attention(Q_{i},K_{i},V_{i}), i \in [1,2,...,h]
\end{aligned}
\end{equation}

\begin{figure*}[ht]
	\centering
% 	\vspace{-20pt}
	\includegraphics[width=1\textwidth]{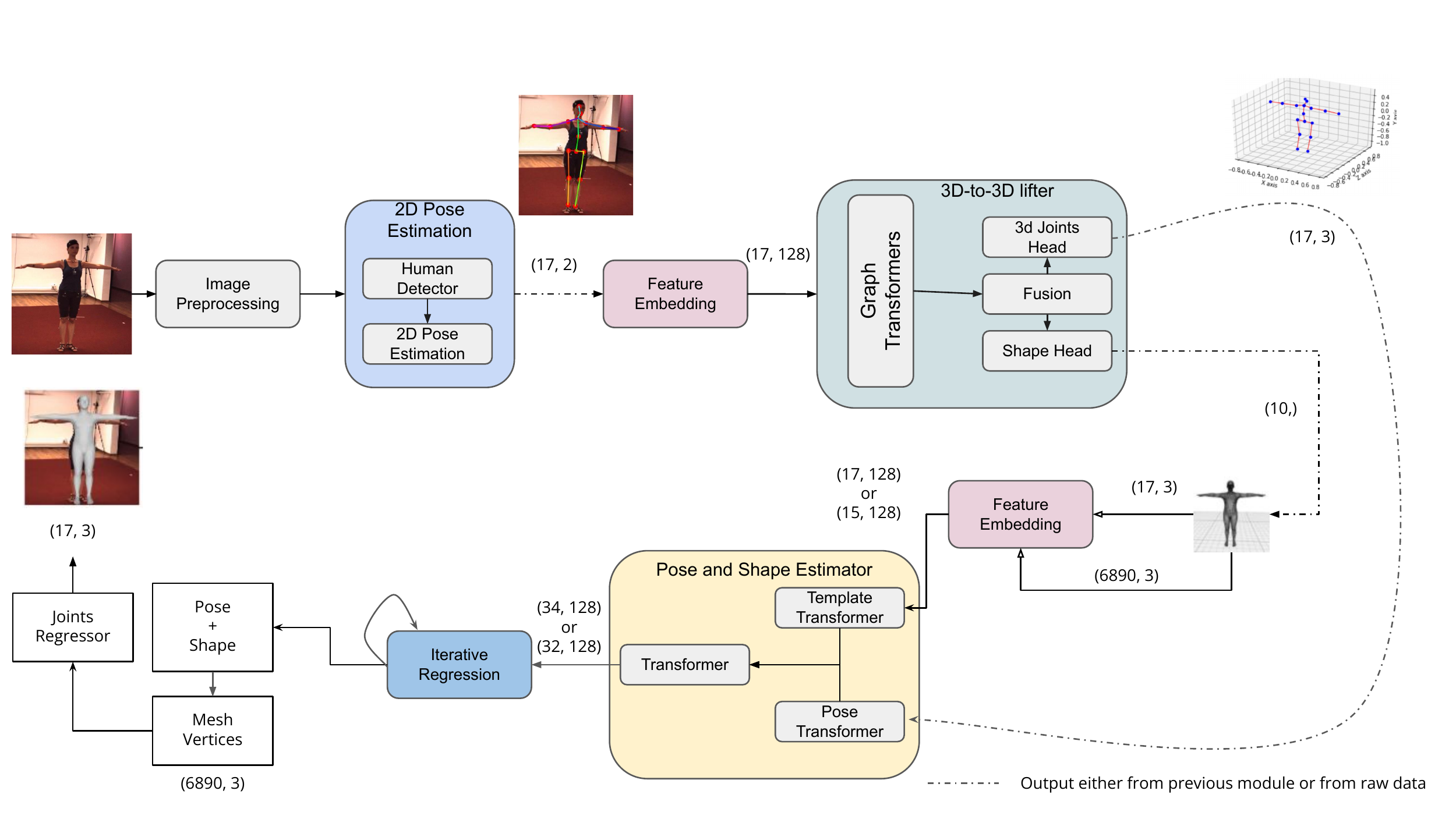}
%  	\vspace{-25pt}
	\caption{\footnotesize{Modular Multi-stage Lightweight Graph Transformer Network
for Human Pose and Shape Estimation from 2D Human Pose}  }
%   \vspace{-12pt}
	\label{fig:multi_stage_pipeline}
\end{figure*}

\subsection{System Architecture}
At first, a projection layer encodes the 2D pose estimation to create latent space embedding to be consumed by the 3D lifter. Supervised training is adopted to train a 3D lifter module where we minimize the distance between the predicted 3D and the ground truth. In addition to using L1 regularization, we use the weak perspective camera model to reproject 3D to the 2D pose to ensure better accuracy and alignment. Furthermore, the shape parameters are applied to the mean template mesh. We experimented with using the mean mesh, which is represented by $M \in \mathbb{R}^{6890}$, and the rest pose regressed from the mean template mesh. We found that using mean mesh provided more gain than using the rest pose. The pose angles are predicated by the iterative regressor that consumes the features generated by the pose and shape estimator. Finally, forward kinematics are employed to calculate the human mesh.

\subsubsection{3D-lifter Module}
The 3D-lifter module leverages the graph transformer as described in \cite{zheng2022lightweight} to extract both global and local features from human pose data. First, a projection layer encodes the pose into a latent space. The resulting features are then fed into multiple parallel graph transformer blocks, which have small embedding dimensions for improved efficiency. The outputs from these parallel branches are fused to generate features that are used to predict 3D pose estimation, human shape, and camera parameters.

Given a 2D skeleton denoted as $S \in \mathbb{R}^{j \times 2}$, where $j$ represents the number of joints and $2$ denotes the joint coordinates in the $X, Y$ plane, the 2D-to-3D lifting task is performed by the Pose Analysis Module (PAM). PAM returns: 1) features embedding $F \in \mathbb{R}^{J \times D}$, where $J$ and $D$ denote the number of joints and the dimensions of the features latent space, respectively; 2) 3D joints $P \in \mathbb{R}^{J \times 3}$, where $J$ denotes the number of joints and $3$ denotes the coordinates of the joints in $[X, Y, Z]$; 3) human shape $\beta \in \mathbb{R}^{10}$, where $10$ represents the first $10$ coefficients of the PCA shape space; and 4) camera parameters $C \in \mathbb{R}^3$, where $3$ denotes the global rotation, translation, and scale.

\subsubsection{Pose and Shape Estimator}
The Pose and Shape Estimator module aims to estimate human mesh parameters, specifically joint angles. System modularity is a crucial goal in this work, and to that end, the module is designed to consume either the 3D-lifter features in an end-to-end training setup or the 3D pose in individual module training. However, due to its lightweight design, the module may not have enough capacity to accurately estimate the pose angle with its small embedding size. To address this, the module is augmented with a mean mesh template as introduced in \cite{zheng2022lightweight} to enrich the feature representation through an additional branch. The features produced by the pose feature and mesh template branches are then fused to generate the final features space, which is subsequently consumed by the iterative 3D regression as described in \cite{kolotouros2019learning} to infer the final 3D pose angle.

Given the pose features on the first branch, expressed as $F_{pose} \in \mathbb{R}^{J \times D}$ where $J$ denotes the number of joints and $D$ denotes the dimensions, and the mesh template features on the second branch, expressed as $F_{template} \in \mathbb{R}^{T \times D}$. Both branches are processed by the same transformer architecture to model the features. The resulting features from both branches are combined into a single final embedding, represented as $F_{PSE} \in \mathbb{R}^{(J + T) \times D}$. Finally, the iterative 3D regression consumes the fused features to generate the pose angles, expressed as $\theta \in \mathbb{R}^{72}$.

\begin{table*}[]
\centering

\begin{tabular}{|c|cc|cc|}
\hline
\multirow{2}{*}{} 
& \multicolumn{2}{c|}{\multirow{2}{*}{Methods}}   & \multicolumn{2}{c|}{Humans.6M}                     \\ \cline{4-5} 
& \multicolumn{2}{c|}{} & \multicolumn{1}{c|}{MPJPE}         & PA-MPJPE      \\ \hline
\multirow{6}{*}{\begin{tabular}[c]{@{}c@{}}Image \\ Based\end{tabular}}              & \multicolumn{1}{c|}{HMR}            & CVPR 2018 & \multicolumn{1}{c|}{88.0}          & 56.8          \\
& \multicolumn{1}{c|}{GraphCMR}       & CVPR 2019 & \multicolumn{1}{c|}{-}             & 50.1          \\
& \multicolumn{1}{c|}{SPIN}           & ICCV 2019 & \multicolumn{1}{c|}{-}             & 41.1          \\
& \multicolumn{1}{c|}{METRO}          & CVPR 2021 & \multicolumn{1}{c|}{54.0}          & 36.7          \\
& \multicolumn{1}{c|}{MeshGraphormer} & ICCV 2021 & \multicolumn{1}{c|}{\textbf{51.2}} & \textbf{34.5} \\
& \multicolumn{1}{c|}{PyMAF}          & ICCV 2021 & \multicolumn{1}{c|}{57.7}          & 40.5          \\ \cline{1-1}  \cline{1-5}
\multirow{2}{*}{\begin{tabular}[c]{@{}c@{}}Video\\ Based\end{tabular}}               & \multicolumn{1}{c|}{VIBE}           & CVPR 2020 & \multicolumn{1}{c|}{65.6}          & 41.4          \\
& \multicolumn{1}{c|}{TCMR}           & CVPR 2021 & \multicolumn{1}{c|}{62.3}          & 41.1          \\ \cline{1-5}
\multirow{3}{*}{\begin{tabular}[c]{@{}c@{}}Pose\\ Based \end{tabular}} & \multicolumn{1}{c|}{Pose2Mesh}      & ECCV 2020 & \multicolumn{1}{c|}{64.9}          & 47.0          \\
& \multicolumn{1}{c|}{Baseline}       & -         & \multicolumn{1}{c|}{68.3}          & 50.0          \\
& \multicolumn{1}{c|}{GTRS}           & -         & \multicolumn{1}{c|}{64.3}          & 45.4          
\\ & \multicolumn{1}{c|}{\textbf{Our}}           & -         & \multicolumn{1}{c|}{65.9}          & 47.13\\   \cline{1-5}
\end{tabular}

	\caption{Our extensive experiments showed our pipeline score comparable results to various contributions }
	\label{fig:local_levelVSserver_level}
\end{table*}

\section{Experimental Evaluation}

\subsection{Datasets}

The \textbf{Human3.6M} dataset \cite{ionescu2013human3} is a widely adopted, video-based, and large-scale indoor dataset that was captured using an accurate marker-based motion-capturing system. The dataset comprises 11 professional actors performing 17 predefined actions. In accordance with previous studies, such as \cite{choi2020pose2mesh} and \cite{kolotouros2019learning}, we selected 5 subjects for training and utilized 2 subjects' samples for evaluation.

The \textbf{3DPW} dataset \cite{von2018recovering} is an outdoor dataset consisting of 60 video sequences that total 51,000 frames captured in an uncontrolled environment. The dataset features ground-truth 3D pose and mesh annotations. In accordance with the evaluation protocol established in previous studies, such as \cite{choi2020pose2mesh} and \cite{kolotouros2019learning}, we adopted the test dataset only.

The \textbf{MuCo-3DHP} dataset \cite{mehta2017monocular} is a large-scale dataset for 3D human pose estimation and mesh reconstruction in natural environments. The dataset includes over 13 million frames of video data recorded by 5 actors performing 7 actions in various outdoor settings. The dataset comprises ground-truth 3D pose and mesh annotations and has been utilized in numerous research studies to evaluate the performance of different algorithms for 3D human pose estimation and mesh reconstruction. Similar to \cite{choi2020pose2mesh} and \cite{kolotouros2019learning}, we utilized this dataset for mixed training.

The \textbf{MSCOCO keypoints} dataset \cite{lin2014microsoft} is a subset of the larger MSCOCO dataset that was created for the purpose of human pose estimation in images. The dataset includes over 200,000 images with ground-truth keypoint annotations for 17 different body joints. The dataset has been widely used in research to evaluate the performance of different algorithms for human pose estimation and has served as a benchmark for comparing the performance of various approaches. In accordance with previous studies, such as \cite{choi2020pose2mesh} and \cite{kolotouros2019learning}, we adopted the MSCOCO dataset for mixed training.

\subsection{Metrics for Evaluation}
The mean per-joint position error (MPJPE) is a commonly used metric in the field of 3D human pose estimation to assess the performance of algorithms. It quantifies the average Euclidean distance between the predicted and ground-truth 3D joint positions in a dataset, after aligning them through a rigid transformation that minimizes the distance between the two sets of points. The MPJPE is expressed in millimeters, with lower values indicating higher accuracy in pose estimation.

The Procrustes aligned mean per-joint position error (PA-MPJPE) is a variant of the MPJPE metric that also evaluates the accuracy of 3D human pose estimation algorithms. Similar to MPJPE, it measures the average Euclidean distance between the predicted and ground-truth joint positions. However, PA-MPJPE incorporates an additional alignment step using the Procrustes analysis method, which aligns the predicted and ground-truth positions by applying a non-rigid transformation that minimizes the overall distance between the two sets of points. The PA-MPJPE is also reported in millimeters, with lower values indicating more accurate pose estimation.

The mean per-vertex error (MPVE) is a metric for evaluating the performance of algorithms for 3D human mesh reconstruction. It quantifies the average Euclidean distance between the predicted and ground-truth 3D mesh vertices in a dataset, after aligning them through a rigid transformation that minimizes the distance between the two sets of points. The MPVE is expressed in millimeters, with lower values indicating higher accuracy in mesh reconstruction.

\section{Conclusion}

In this work, we present a pose-based framework for reconstructing human mesh parameters from 2D poses. This framework utilizes a 3D-lifter that leverages structured and implicit joint correlations through the employment of paralleled graph transformer blocks. As a result, the model is able to efficiently integrate the extracted pose features with the mesh template to produce human mesh parameters. Although the proposed framework demonstrates competitive performance, it may not be capable of reconstructing all human body shapes solely from 2D pose inputs. While image-based methods may achieve more precise reconstructions, pose-based approaches maintain their significance owing to their versatility and lightweight design. Additionally, the modular design of the proposed framework allows for training to be conducted in individual or end-to-end modules.

\bibliography{main}

\end{document}